\documentclass[10pt,twocolumn,letterpaper]{article}

\usepackage{cvpr}
\usepackage{times}
\usepackage{epsfig}
\usepackage{graphicx}
\usepackage{amsmath}
\usepackage{amssymb}
\usepackage{graphics}
\usepackage[table,xcdraw]{xcolor}

\newcommand{\Rd}{{\mathbb R}}

\newcommand{\Xc}{{\mathcal X}}
\newcommand{\Yc}{{\mathcal Y}}
\newcommand{\Fc}{{\mathcal F}}

\usepackage{amsmath,amssymb,nopageno}
\usepackage[all]{xy}

\usepackage[pagebackref=true,breaklinks=true,letterpaper=true,colorlinks,bookmarks=false]{hyperref}

\cvprfinalcopy % *** Uncomment this line for the final submission

\def\cvprPaperID{29} % *** Enter the CVPR Paper ID here

\ifcvprfinal\fi
\begin{document}

%%%%%%%%% TITLE
\title{Beyond Deep Residual Learning for Image Restoration: \\ Persistent Homology-Guided Manifold Simplification }

\author{Woong Bae*, Jaejun Yoo*, and Jong Chul Ye\\
Bio Imaging Signal Processing Lab\\
Korea Ad. Inst. of Science \& Technology (KAIST) \\
291 Daehak-ro, Yuseong-gu, Daejeon 34141, Korea
\\
{\tt\small \{iorism,jaejun2004,jong.ye\}@kaist.ac.kr}\\
{\small * denotes co-first authors}% For a paper whose authors are all at the same institution,
}

\maketitle
%\thispagestyle{empty}

%%%%%%%%% ABSTRACT
\begin{abstract}
The latest deep learning %based image restoration
approaches perform better than the state-of-the-art signal processing approaches in various
image restoration tasks. However, if an image contains many patterns and structures, the performance of these CNNs is still inferior. % to BM3D.
To address this issue, here we propose a novel feature space deep residual learning algorithm that outperforms the existing residual learning. The main idea is originated from the observation
that the performance of a learning algorithm can be improved if the input and/or label manifolds can
be made topologically simpler by an analytic mapping to a feature space. 
%Using persistent homology analysis,
% we show that the recent residual learning was benefited from such manifold simplification.
Our extensive numerical studies using denoising experiments and NTIRE single-image super-resolution (SISR) competition demonstrate that the proposed feature space residual learning
outperforms the existing state-of-the-art approaches. Moreover, our algorithm was ranked third in NTIRE competition with 5-10 times faster computational time compared to the top ranked teams. The source code is available on page : \href{url}{https://github.com/iorism/CNN.git}

\end{abstract}

%%%%%%%%% BODY TEXT
\section{Introduction}

Image restoration tasks such as denoising and super-resolution are essential steps in many practical image processing applications. %, so they has been extensively studied. 
Over the last few decades, various algorithms have been developed, which include
non-local self-similarity (NSS) models \cite{buades2008nonlocal}, total variation (TV) approaches \cite{osher2005iterative}, and sparse dictionary learning models \cite{dong2013nonlocally}. Among them, 
the block matching 3D filter (BM3D) \cite{dabov2007image} % and weighted nuclear norm minimization (WNNM) \cite{gu2014weighted} 
is considered
as the state-of-the art algorithm.
In general, these methods are dependent on the noise model.
Moreover, these algorithms are usually implemented in an iterative manner, so they require significant computational resources.

Recently, deep learning approaches have achieved tremendous
success in classification~\cite{krizhevsky2012imagenet} as well as low-level computer vision problems ~\cite{ronneberger2015u}. 
%Theoretical origin of their success has been investigated by a few authors \cite{poole2016exponential,telgarsky2016benefits}, where
%the exponential expressivity under a given network complexity (e.g. VC dimension \cite{anthony2009neural} or Rademacher complexity \cite{bartlett2002rademacher})
%has been attributed to their success.
%
%
In image denoising and super-resolution tasks, many state-of-the-art CNN algorithms \cite{chen2015trainable,chen2015learning,mao2016image,zhang2016beyond} have been proposed.
Although the performance of these algorithms usually outperforms the non-local and collaboration filtering approaches such as BM3D,
in case of certain images that have many patterns (such as Barbara image), CNN approaches are still inferior to BM3D.

Therefore, one of the main motivations of this work is to develop a new CNN architecture that 
overcomes the limitation of the state-of-the-art CNN approaches. % and non-local and collaboration filtering approaches such as BM3D.
%In particular, we propose a novel {\em wavelet-domain deep residual learning} to learn the input/label mappings in the wavelet domain.
%Once the wavelet domain residuals are estimated, then the final noise-free or super-resolution image can be obtained by subtracting the estimated residual as shown in Fig.~\ref{fig:Proposed_Network_base}.
%
The proposed network architectures are motivated from a novel {\em persistent homology} analysis \cite{edelsbrunner2008persistent} on residual learning for image processing tasks. %\cite{zhang2016beyond}.
Specifically, we show that the residual manifold is 
topologically simpler than the original image manifold, which may have attributed the success of residual learning. 
Moreover, this observation leads us to a new network design principle using {\em manifold simplification}.
Specifically, our design goal is to find mappings for input and/or label datasets to feature spaces, respectively, such that
the new datasets become topologically simpler and easier to learn.
In particular, %using a recent computational topology tool called the
we show that the wavelet transform provides topologically simpler manifold structures while
preserving the directional edge information.

\begin{figure*}[!hbt]
\centerline{\includegraphics[width=0.9\linewidth]{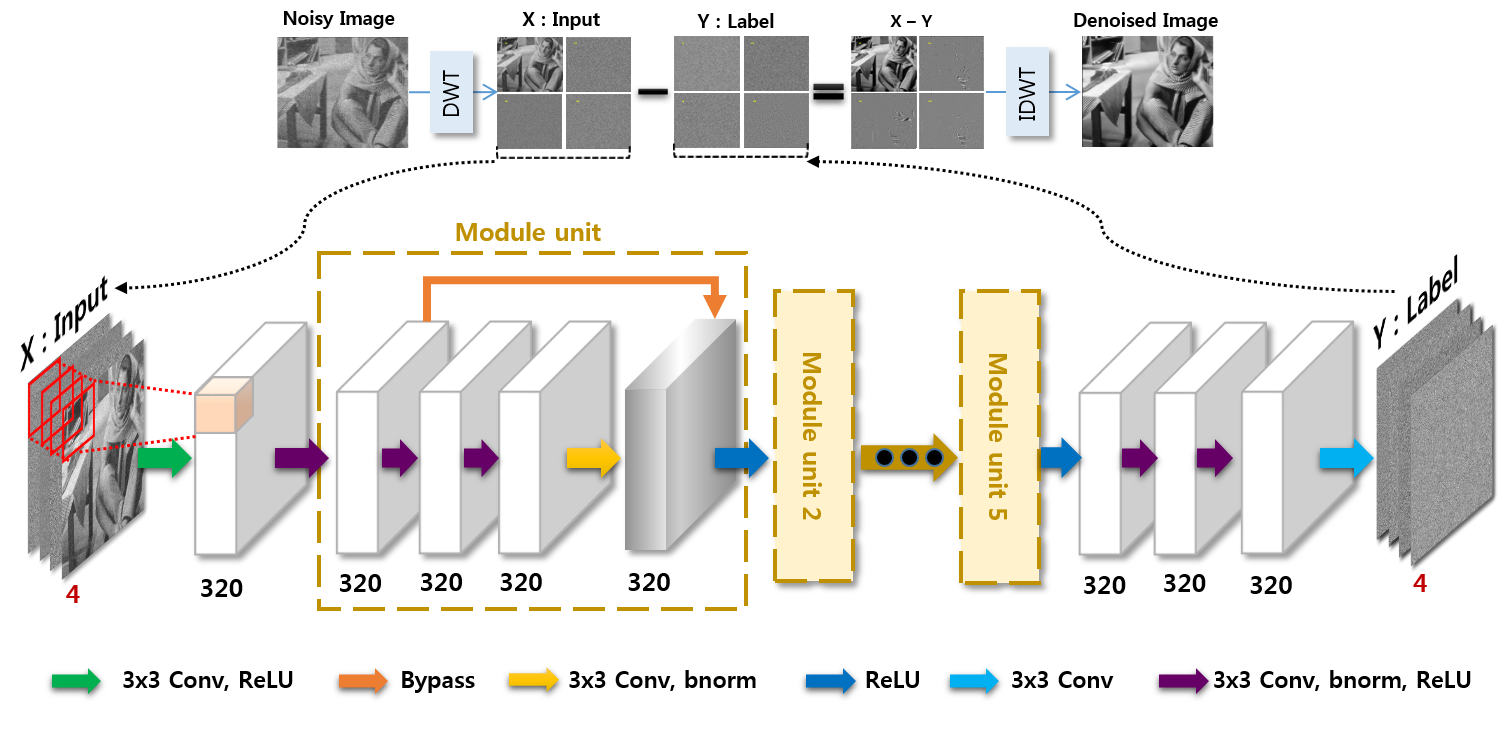}}
\caption{ Proposed wavelet domain deep residual learning network for the Gaussian denoising task.} %$^*$Yellow arrow ($3\times3$ Conv, bnorm, ReLU) is excluded for between adjacent matrices.
\label{fig:Proposed_Network_base}
\end{figure*}

\begin{table*}[!hbt]
\begin{center}
%\resizebox{\columnwidth}{!}{%
\scalebox{0.75}{
\begin{tabular}{|c|c|c|c|c|} 
\hline
%\multirow{4}{1}{XXX}
~& Bicubic x2 (256ch) & Bicubic x3,x4 (320ch)
& Unknown x2,x3,x4 (320ch) \\ %[0.5ex] 
\hline
\rowcolor[HTML]{C0C0C0} 
Input & \multicolumn{2}{|c|}{ WT( BU(LR) )} & COPY\_ch(LR) \\ %[0.5ex] 
\hline
\rowcolor[HTML]{C0C0C0} 
Label & \multicolumn{2}{|c|}{ Input - WT(HR)}& Input - PS(HR) \\ %[0.5ex] 
\hline 
1st layer & \multicolumn{2}{c|}{ Conv $\to$ ReLU} & Conv $\to$ ReLU\\ %[0.5ex] 
\hline 
2nd layer & \multicolumn{2}{c|}{ Conv $\to$ BN $\to$ ReLU} & Conv $\to$ BN $\to$ ReLU\\ %[0.5ex] 
\hline 
Long bypass layer & LongBypass(1) &- &-\\ %[0.5ex] 
\hline 
1st module &\multicolumn{1}{c|}{\begin{tabular}[c]{@{}c@{}}BypassM1 $\to$\\ Conv$\to$BN$\to$ReLU$\to$\\ Conv$\to$BN$\to$ReLU$\to$\\ Conv $\to$BN$\to$\\ SumF(BypassM) $\to$\\ ReLU \end{tabular} }&\multicolumn{1}{c|}{\begin{tabular}[c]{@{}c@{}}BypassM1 $\to$\\ Conv$\to$BN$\to$ReLU$\to$\\ Conv$\to$BN$\to$ReLU$\to$\\ Conv $\to$BN$\to$\\ SumF(BypassM) $\to$\\ ReLU \end{tabular} }&\multicolumn{1}{c|}{\begin{tabular}[c]{@{}c@{}}BypassM1 $\to$\\ Conv$\to$BN$\to$ReLU$\to$\\ Conv$\to$BN$\to$ReLU$\to$\\ Conv $\to$BN$\to$\\ SumF(BypassM) $\to$\\ ReLU \end{tabular} }\\ %[0.5ex] 
\hline 
Repeat 1st module & 5 times (2th$\sim$6th module) & 11 times (2th$\sim$12th module) & 12 times (2th$\sim$12th module) \\ %[0.5ex] 
\hline 
Long bypass \& catch layer & \multicolumn{1}{c|}{\begin{tabular}[c]{@{}c@{}}Sum of ``LongBypass(1)'' and \\``Output of 6th module''\\$\to$ BN$\to$ ReLU\end{tabular} } & - &- \\ %[0.5ex] 
\hline 
Long bypass layer & LongBypass(2) &- &-\\ %[0.5ex] 
\hline 
Repeat 1st module & 6 times (7th$\sim$12th modules) &- & - \\ %[0.5ex] 
\hline 
Long bypass \& catch layer &\multicolumn{1}{c|}{\begin{tabular}[c]{@{}c@{}}Sum of ``LongBypass(2)'' and \\``Output of 12th module''\\$\to$ BN$\to$ ReLU\end{tabular} } & -&-\\ %[0.5ex] 
\hline 
Last layer & \multicolumn{1}{c|}{\begin{tabular}[c]{@{}c@{}}Conv$\to$BN$\to$ReLU$\to$\\ Conv$\to$BN$\to$ReLU$\to$\\ Conv \end{tabular} } & \multicolumn{1}{c|}{\begin{tabular}[c]{@{}c@{}}Conv$\to$BN$\to$ReLU$\to$\\ Conv$\to$BN$\to$ReLU$\to$\\ Conv \end{tabular} } & \multicolumn{1}{c|}{\begin{tabular}[c]{@{}c@{}}Conv$\to$BN$\to$ReLU$\to$\\ Conv$\to$BN$\to$ReLU$\to$\\ Conv \end{tabular} } \\ %[0.5ex] 
\hline 
\rowcolor[HTML]{C0C0C0} 
Restoration &\multicolumn{2}{c|}{IWT(Input-Output)} & IPS(Input-Output)\\ %[0.5ex] 
\hline 
\multicolumn{4}{p{1.2\textwidth}}{* {\bf WT}: Haar Wavelet Transform, {\bf BU}: Bicubic Upsampling, {\bf LR}: Low Resolution image, {\bf HR}: High Resolution image, {\bf Conv}: $3\times3$ Convolution, {\bf BN}: Batch Normalization, {\bf BypassM}: Sending output of previous layer to last layer of module(M is module number), {\bf SumF}: Sum of output of previous layer and BypassM output, {\bf COPY\_ch}: Copy input image (scale x scale) times on channel direction, {\bf PS}: sub-Pixel Shuffling, {\bf IPS}: Inverse sub-Pixel Shuffling, {\bf IWT}: Inverse Wavelet Transform} \\ %[0.5ex] 

\end{tabular}
}
\end{center}
\caption{Proposed network architectures for NTIRE SISR competition from bicubic and unknown downsampling schemes.}
\label{tb:Proposed_Network_NTIRE}
\end{table*}

{\bf Contribution}: In summary, our contributions are as following. First, a novel network design principle using manifold simplification
is proposed. % by introducing invertible linear mappings to input and/or label.
Second, using a recent
computational topology tool called the persistent homology, 
we show that the existing residual learning is a special case of manifold simplification
and then propose a wavelet transform to simplify topological structures of input and/or label manifolds.

%-------------------------------------------------------------------------
\section{Related Work}

One of the classical approaches for image denoising is a wavelet shrinkage approach \cite{donoho1995noising}, which decomposes an image into low and high frequency subbands 
and applies thresholding in the high frequency coefficients \cite{portilla2003image}.
Advanced algorithms in this field are to exploit the intra- and inter- correlations of the wavelet coefficients \cite{crouse1998wavelet}.

In neural network literature, the work by Berger et al \cite{burger2012image} was the first which demonstrated similar denoising performance to BM3D using multi-layer perceptron (MLP). 
Chen \etal \cite{chen2015trainable,chen2015learning} proposed a deep learning approach called trainable
nonlinear reaction diffusion (TNRD) that
can train filters and influence functions by unfolding a variational optimization approach. %The approach was extended to deal with image super-resolution and JPEG deblocking \cite{chen2015trainable}.
Recently, based on skipped connection and encoder-decoder architecture, a very deep residual
encoder-decoder networks (RED-Net) was proposed for image restoration problems \cite{mao2016image}.

Residual learning has multiple realizations. The first approach is using a skipped connection that bypasses input data of a certain layer to another layer during forward and backward propagations.
This type of residual learning concept was first introduced by He \etal\cite{he2015deep} for image recognition.
In low-level computer vision problems, Kim \etal\cite{kim2015accurate} employed a residual learning for a super-resolution method. 
In these approaches, the residual learning was implemented by a skipped connection corresponding to an identity mapping.
In another implementation, the label data is transformed into the difference between the input data and clean data. 
For example, Zhang \etal\cite{zhang2016beyond} proposed a denoising convolutional neural networks (DnCNNs) \cite{zhang2016beyond}, which has inspired our method.

\section{Theory}
%Before we explain the proposed architecture, this section provides
%some theoretical backgrounds. %h implementation.

\subsection{Generalization bound}

Let 
$X \in \Xc$ and $Y \in \Yc$ denote the input and label data and $f: X \rightarrow Y$ denotes a function living in a functional space $\Fc$.
Then, one is interested in the minimization problem:
%that minimizes the risk
%\begin{eqnarray}
$\min_{f \in \Fc} L(f)$, 
%\end{eqnarray}
where
$L(f) = E_D \|Y-f(X)\|^2$ denotes the risk.
A major technical issue %n minimizing $L(f)$
is, however, that the associated probability distribution $D$ is unknown.
Thus, an upper bound is used to characterize the generalization performance.
Specifically, with probability $\geq 1-\delta$ with a small $\delta>0$, for every function $f\in \Fc$, %we have
\begin{equation}\label{eq:L}
L(f) \leq \underbrace{\hat L_n(f)}_{\text{empirical risk}} + \underbrace{2 \hat R_n(\Fc)}_{\text{complexity penalty}}+ 3 \sqrt{\frac{\ln(2/\delta)}{n}}
\end{equation}
where $\hat R_n(\Fc)$ denotes the Rademacher complexity \cite{bartlett2002rademacher}.
%is defined to be
%$$\hat R_n(\Fc)= E_\sigma \left[\sup_{f\in \Fc} \left(\frac{1}{n}\sum_{i=1}^n \sigma_i f(X_i) \right) \right],$$ 
%where $\sigma_1,\cdots, \sigma_n$ are independent random variables uniformly chosen from $\{-1,1\}$.
%In general, this complexity term is determined by
%the structure of a network. 
%Therefore, to reduce the risk, we need to minimize both the empirical risk and the complexity terms in \eqref{eq:L} simultaneously.

In neural network, empirical risk is determined by the representation power or capacity of a network \cite{telgarsky2016benefits},
and the complexity penalty is determined by the structure of a network.
It was shown that the capacity of representation power grows exponentially with respect to the number of layers \cite{telgarsky2016benefits}, which justifies the use of a {\em deep } network compared to shallow ones.
However, the complexity penalty in \eqref{eq:L} also increases with a complicated network structure.
The main remedy for this trade-off is to use large number of training dataset such that the contribution of the complexity penalty
reduces much more quickly so that the empirical risk minimization (ERM) converges consistently to the risk minimization \cite{vapnik1998statistical}.

However, for the intermediate size of the training data, there still exist gaps between the ERM and the risk minimization.
One of the most important contributions of this paper is to reduce the gap by using a relatively simpler network,
by reducing the complexity of the data manifold. % such that the simple network can still fit the data with small empirical error.

Specifically, for a given deep network $f: X\rightarrow Y$,
our design goal is to find mappings $\Phi$ and $\Psi$ to the {\em feature spaces} for the input and label datasets, respectively. Then, the resulting datasets composed of $X'=\Phi(X)$ and $Y'=\Psi(Y)$  may have simpler manifold structures.
This can be shown in the following diagram:
\begin{eqnarray*}
\xymatrix@C+1em@R+1em{
X \ar[r]^{f}
\ar@<-2pt>[d]_{\Phi} & Y \ar@<-2pt>[d]_{\Psi} \\
X'\ar@<-2pt>[u]_{\Phi^{-1}} \ar[r]^{g} & Y' \ar@<-2pt>[u]_{\Psi^{-1}}
}
\end{eqnarray*}
from which our goal is to find an equivalent neural network $g: X' \rightarrow Y'$ that has a better performance
than the original network $f: X \rightarrow Y$.

For example, in recent deep residual learning \cite{zhang2016beyond}, the input transform $T$ is an identity mapping
and the label transform is given by %relationship:
%\begin{eqnarray*}
$Y'=\Psi(Y) = Y-X$ .
%\end{eqnarray*}
%whose inverse transform is given by
%$$ S^{-1}(Y')= X+Y' .$$
Using persistent homology analysis, Section 1 in the supplementary material shows that the label manifold of the residual is topologically simpler than that of $Y$.
%We maintain that this is the main reason that residual network is usually better. 
Accordingly, the upper bound of the risk of $g: X' \rightarrow Y'$ can be reduced compared to that of $f: X\rightarrow Y$.

Inspired by this finding, this paper proposes a wavelet transform as 
a good transform to reduce the topological complexity of resulting input and label manifolds.
More specifically, thanks to the vanishing moments of wavelets, the wavelet transform can annihilate the 
smoothly varying signals while retaining the image edges \cite{daubechies1992ten,mallat1999wavelet}, which results in the dimensional reduction and
manifold simplification.
% Indeed, this results in dimensional reduction. 
Indeed, this property of the wavelet transform 
has been extensively exploited in wavelet-based image compression
tools such as JPEG2000 \cite{skodras2001jpeg}, and this paper shows that this property 
also improves the performance of deep network for image restoration tasks.

\begin{figure}[!bt]
\centerline{\includegraphics[width=0.99\linewidth]{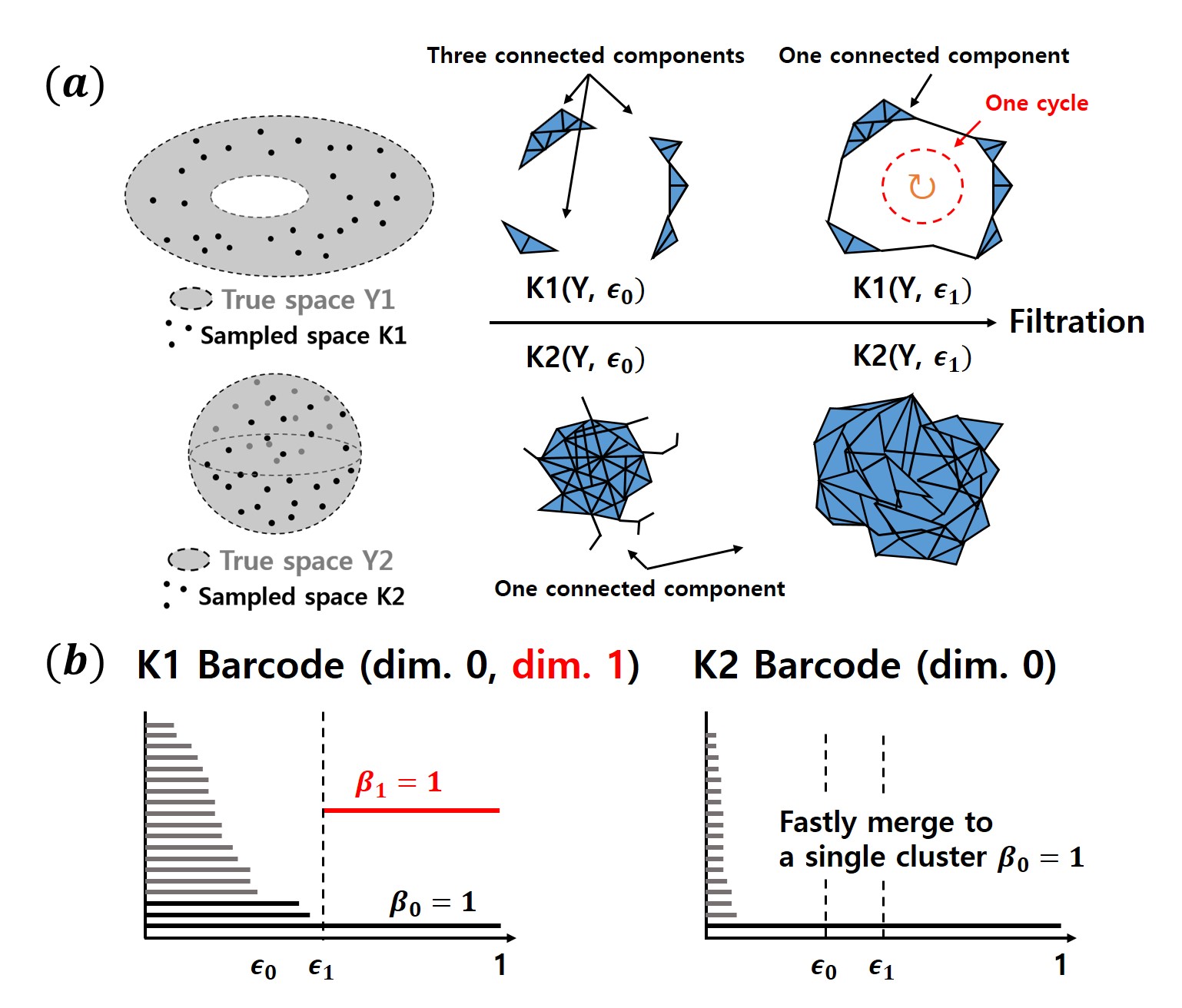}}
\caption{(a) Point cloud data $K$ of true space $Y$ and its configuration over $\epsilon$ distance filtration. $Y1$ is a doughnut and $Y2$ is a sphere shaped space each of which represents a complicated space and a simpler space, respectively. (b) Zero and one dimensional barcodes of $K1$ and $K2$. Betti number can be easily calculated by counting the number of barcodes at each filtration value $\epsilon$. 
%(c) An exemplary process of generating persistence landscape from barcodes. Persistence landscape can be easily obtained by transforming each barcode $\{(b_i,d_i)\}_{i=1}^3$ in a triangular shape with height of $\frac{d_i-b_i}{2}$. Here, $\lambda_1$ is the outmost landscape and $\lambda_2$ is the second outmost landscape. %Note that $\lambda_1$ gives a measure of the dominant feature at each $\epsilon$.
}
\label{fig:topo_exp}
\end{figure}

\subsection{Persistent homology}\label{sec:ph}

The complexity of a manifold is a topological concept. Thus, it should be analyzed using topological tools.
In algebraic topology, Betti numbers ($\beta_m$) represent the number of $m$-dimensional holes of a manifold. For example, $\beta_0$ and $\beta_1$ are the number of connected components and cycles, respectively. They are frequently used to investigate the characteristics of underlying manifolds \cite{edelsbrunner2008persistent}. 
Specifically, we can infer the topology of our data manifold by varying the similarity measure between the data points and tracking the changes of Betti numbers. % distance filtration.
As allowable distance $\epsilon$ increases, data point clouds merge together and finally become a single cluster (Fig.~\ref{fig:topo_exp}(a)). 
Therefore, the point clouds with high diversity will merge slowly and this will be represented as a slow decrease in Betti numbers. 
For example, in Fig.~\ref{fig:topo_exp}(a), the dataset $Y1$ is a doughnut with a hole (i.e. $\beta_0=1$ and $\beta_1=1$) whereas
$Y2$ is a sphere-like cluster (i.e. $\beta_0=1$ and $\beta_1=0$). 
Accordingly, $Y1$ has longer zero dimensional \emph{barcodes} persisting over $\epsilon$ in Fig.~\ref{fig:topo_exp}(b). 
%In other words, $Y1$ data set has a 
% distanced configuration of point clouds that cannot be overcome until they reach a large $\epsilon$.
This persistence of Betti number is an important topological feature and the
recent {\em persistent homology} analysis utilizes this to investigate the topology of the data manifold\cite{edelsbrunner2008persistent}.

%{
%For quantitative comparison of the differences in topology, a statistical testing technique using the topological summary called {\em persistence landscape} has been developed \cite{bubenik2015statistical}. In the standard paradigm for topological data analysis, the end result is a topological summary of the data, e.g. barcodes. Persistence landscape is 
%a conversion of barcodes such that
%it has a good notion of mean.
%Specifically, for a given set of barcodes $\{ (b_i,d_i)\}_{i=1}^n$, the $k$-th landscape
%$ \lambda_k(t)$ is defined as the $k$-th largest value of $\min(t-b_i,d_i-t)_+,$
%where $b_i$ and $d_i$ stand for the birth and death points of the barcode and $(\cdot)_+$ denotes $\max(\cdot,0)$ (Fig.\ref{fig:topo_exp}(c)). 
%It is proven to obey a strong law of large numbers and a central limit theorem so that we can directly use the standard statistical tests for the inference \cite{bubenik2015statistical}.
%%There exist maps in both directions between barcode and persistence landscape.
%Considering the landscape value
%%\begin{equation}\label{eq:lambda}
%$|| \pmb{\Lambda} ||_1=\sum_{k=1}^{\infty}||\lambda_k||_1$
%%\end{equation}
%as a real random variable, we may use the normal distribution to calculate its confidence interval and perform hypothesis testing whether
%the two manifolds are topologically different or not \cite{bubenik2015statistical}. 
%}

\section{Proposed architecture}
%{\bf THIS NEEDS TO BE RE-WRITTEN. }
This section describes two network structures based on the manifold simplificaiton. One is the primary architecture used for Gaussian denoising and the other is our NTIRE 2017 competition architecture used for RGB based SISR problems, which has been developed based on the primary architecture. For the wavelet transform, we used one level discrete wavelet transform using Haar wavelet filter.

{\bf Denoising architecture}: %As shown in Fig.\ref{fig:Proposed_Network_base}, 
The input and the clean label images are first decomposed into four subbands (i.e. LL, LH, HL, and HH) using the wavelet transform. 
The wavelet residual images, which are now used as our new labels, are obtained by the difference between the input and the clean label images in the wavelet domain. Then, the network is trained to learn multi-input and multi-output functional relationship between these newly processed input and label.
%As will be shown later in persistent homology analysis in Appendix, this wavelet transform reduces the manifold complexity of inputs and labels.
Here, four patches at the same locations in each wavelet subband are extracted and used for training.
For Gaussian denoising, $40\times 40$ image patches are used, resulting in $40\times 40\times 4$ patches. 
%For super-resolution,
%$20\times 20\times 4$ patches were used.
%Thanks to the shift invariance, for testing, whole wavelet transform subbands are used rather than processing in a patch-by-patch manner.

The proposed denoising network architecture is shown in Fig.~\ref{fig:Proposed_Network_base}. 
%It is similar to the identity mapping network in \cite{he2015deep}. % The major differences Differences are count of skipped convolution module and discard batch normalization at first convolution module. 
It consists of five modules between the first and the last stages. Each module has one bypass connection, three convolution layers, three batch normalizations \cite{ioffe2015batch}, and three Rectified Linear Unit (ReLU) \cite{glorot2011deep} layers. 
The bypass connection was used for an efficient network training because it is helpful for training a deep network by alleviating the gradient vanishing problem \cite{he2016identity,mao2016image}. 
The first stage contains two layers: one with a convolution layer with ReLU which is followed by the other convolution layer with batch normalization and ReLU.
%Unlike \cite{he2015deep}, batch normalization was not applied for the first layer, because the wavelet transform makes high frequency band to zero mean.
The last stage is composed of three layers: two layers with a convolution, batch normalization, and ReLU and the last layer with a convolution layer.
Accordingly, the total number of convolution layers is 20.
% whereas the last layer is composed of two convolution layers with ReLu.
The convolution filter size is $3\times 3\times 320 \times 320$.
% , which corresponds to
%$3\times 3$ filters across 320 input and output channels. 
%{\bf
During the convolution, we used zero padding to maintain the image size and reduce the boundary effect \cite{kim2015accurate}.

\begin{figure}[t]
\centerline{\includegraphics[width=0.7\linewidth]{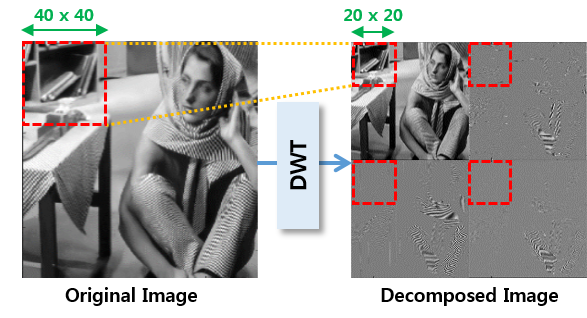}}
\caption{Wavelet decomposition reduces the patch size to a quater.}
\label{fig:Advantages_wavelet1}
\end{figure}

%{\bf
In addition to the aforementioned advantage of the wavelet transform for feature space mapping, there are two more advantages to perform the wavelet transform. As shown in Fig.\ref{fig:Advantages_wavelet1}, the first advantage is that the patch size can be reduced by half. It can reduce the runtime of the network due to the size of the output images of layers being halved. The second one is that the minimum required size of receptive field can be reduced to obtain a good performance. %There is a minimum required receptive field size for image reconstruction using deep running. Therefore, 
Since the number of convolutions is related to the runtime and learning time, the smaller the required receptive field size, the more effective it is to reduce the computation time.

{\bf NTIRE SISR competition architecture}: 
The proposed networks for NTIRE 2017 SISR competition are shown in Table~\ref{tb:Proposed_Network_NTIRE}. These architectures are extended from the primary denoising architecture.
Depending on the decimation schemes (bicubic x2, x3, x4, and unknown x2, x3, and x4) for low resolution dataset, we implemented three different architectures.

\begin{figure}[!h]
\centerline{\includegraphics[width=0.83\linewidth]{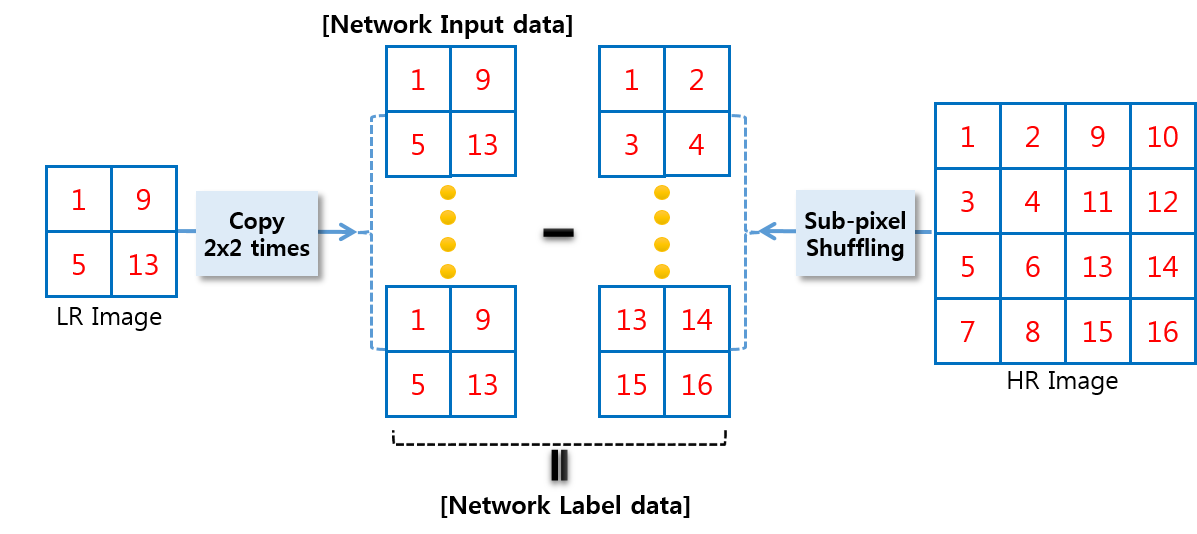}}
\caption{Residual based sub-pixel shuffling.}
\label{fig:residual_subpixelshuffling}
\end{figure}

Specifically, for the bicubic cases, we first generated the upsampled image using bicubic interpolation and the extended denoising network structures with the wavelet transform were used for manifold simplification.
For the unknown decimation scheme, however, we employed the sub-pixel shuffling scheme \cite{shi2016real} as the input and label transform to save the memory and augment the input data to a bigger image size. 
As will be shown later in persistent homology analysis in the supplementary material, this sub-pixel shuffling transform does not reduce the manifold complexity by itself. Still, we could exploit the manifold simplification from the residual learning in sub-pixel shuffling domain as shown in Fig.~\ref{fig:residual_subpixelshuffling} and Table \ref{tb:unknownx3_ori_vs_res_ep50} .

\begin{table}[!h]
\begin{center}
%\resizebox{\columnwidth}{!}{%
\scalebox{0.83}{
\begin{tabular}{|c|c|c|c|c|c|} 
\hline
Label image set & Original & Residual \\ %[0.5ex] 
\hline\hline
Unknown x3 & 29.1242 / 0.8327 & {\bf 30.3025 / 0.8572} \\ %[0.5ex] 
%\hline 
%512x512 & 1.769 & 0.135 \\ %[0.5ex] 
\hline 

\end{tabular}
}
\end{center}
\caption{The effectiveness of the residual based sub-pixel shuffling in terms of PSNR/SSIM for ``Unknown x3' datset in the super-resolution task. The training step was stopped at epoch 50 and the results are calculated from 100 validation data of DIV2K dataset. }
\label{tb:unknownx3_ori_vs_res_ep50}
\end{table}

All three SISR architectures have 41 convolution layers to deal with the dataset composed of 800 high resolution images. In every case, we used $20\times 20$ patch size.
After the first two layers, a basic module is repeated twelve times, which is followed by three convolution layers.
To reconstruct the bicubic x2 downsampled dataset, we included two long bypass connections between six basic modules in the network and the number of channels were 256. For the other datasets, we did not use the long bypass connection and the number of channels were 320.
%{\bf 
The long bypass connection allows faster computation and less parameter size than using the concatenation layer. Although concatenation layer is good for reducing the depth of convolution layer, it is very slow because of inefficient GPU memory usage. %Of course, running by the patch can solve this problem, but it increases redundancy calculation and eventually increases the runtime. 
Thus, the long bypass connection is more efficient for SISR problem. Table ~\ref{tb:longBypassCompare} shows the effectiveness of the long bypass connection.
%{\bf 

We used RGB data for different channels rather than luminance channel, because RGB based learning has the effect of data augmentation.
% because it creates more various features than using the luminance only. Furthermore, the error of domain conversion between YCbCr and RGB deteriorates the results.

\begin{table}[!h]
\begin{center}
%\resizebox{\columnwidth}{!}{%
\scalebox{0.83}{
\begin{tabular}{|c|c|c|c|c|c|} 
\hline
Problem & Without LongBypass & LongBypass \\ %[0.5ex] 
\hline\hline
Bicubic x2 & 35.3436 / 0.9426 & {\bf 35.3595 / 0.9427} \\ %[0.5ex] 
%\hline 
%512x512 & 1.769 & 0.135 \\ %[0.5ex] 
\hline 

\end{tabular}
}
\end{center}
\caption{The effectiveness of the long bypass layer in terms of PSNR/SSIM. This result was calculated from 50 validation data of DIV2K dataset.}
\label{tb:longBypassCompare}
\end{table}

\section{Methods}
\label{sec:exp}

\subsection{Dataset}
{\bf Dataset for denoising network}:
We used publicly available Berkeley segmentation (BSD500) \cite{chen2015trainable} and Urban100 \cite{huang2015single} datasets. 
%These datasets have been commonly used in training and evaluating image restoration networks. 
%%\textbf{ 
Specifically, we used 400 images of BSD500 and Urban100 datasets for training in the Gaussian denoising task. In addition, we generated 4000 training images by using data augmentation via image flipping, rotation, and cropping. To get various noise patterns and avoid overfitting, we re-generated the Gaussian noise in every other epoch during training. For the test dataset, BSD68 and Set12 datasets were used. 
All the images were encoded with eight bits, so the pixel values are within [0, 255].
For training and validation, Gaussian noises with $\sigma=15,~30$, and $50$ were added.

\begin{table*}[!hbt]
\begin{center}
%\resizebox{\columnwidth}{!}{% 0.7
\scalebox{0.7}{
\begin{tabular}{|c|c|c|c|c|c|c|c|c|c|c|c|c|c|} 
\hline
Images & C.man & House &Peppers & Starfish & Monar. & Airpl. & Parrot & Lena & Barbara & Boat & Man & Couple & Average \\ %[0.5ex] 
\hline\hline
Algorithm & \multicolumn{13}{c|}{ Noise Level: $\sigma$ = 30} \\ 
\hline
BM3D & 28.6376 & 32.1417 & 29.2140 & 27.6354 & 28.3458 & 27.4857 & 28.0707 & 31.2388 & 29.7894 & 29.0465 & 28.8016 & 28.8417 & 29.1041 \\
% \hline
% WNNM & 28.7827 & 32.6624 & 29.5032 & 28.1110 & 28.9613 & 27.7509 & 28.2525 & 31.4505 & \bf 30.2572 & 29.1412 & 28.9238 & 28.9389 & 29.3947 \\
\hline 
DnCNN-S & 29.2748 & 32.3199 & 29.8497 & 28.3970 & 29.3165 & 28.1570 & 28.5375 & 31.6104 & 28.8925 & 29.3117 & 29.2492 & 29.2091 & 29.5105 \\
\hline
Proposed(Primary) & \bf 29.6219 & \bf 32.9357 & \bf 30.1054 & \bf 29.0584 & \bf 29.5597 & \bf 28.3288 & \bf 28.6770 & \bf 32.0163 & \bf 29.8941 & \bf 29.6107 & \bf 29.4065 & \bf 29.5563 & \bf 29.8976 \\
\hline
\end{tabular}
}
\end{center}
\caption{Performance comparison in terms of PSNR for ``Set12'' dataset in the Gaussian denoising task. The primary architecture was used.}
\label{Set12_PSNR_denoising}
%\end{table*}
%
%\begin{table*}[!h]
\begin{center}
%\resizebox{\columnwidth}{!}{%
\scalebox{0.7}{
\begin{tabular}{|c|c|c|c|c|c|c|c|c|c|c|c|c|c|} 
\hline
Images & C.man & House &Peppers & Starfish & Monar. & Airpl. & Parrot & Lena & Barbara & Boat & Man & Couple & Average \\ %[0.5ex] 
\hline\hline
Algorithm & \multicolumn{13}{c|}{ Noise Level: $\sigma$ = 30} \\ 
\hline
BM3D & 0.8373 & 0.8480 & 0.8502 & 0.8282 & 0.8866 & 0.8361 & 0.8320 & 0.8456 & 0.8673 & 0.7777 & 0.7783 & 0.7937 & 0.8318 \\
% \hline
% WNNM & 28.7827 & 32.6624 & 29.5032 & 28.1110 & 28.9613 & 27.7509 & 28.2525 & 31.4505 & \bf 30.2572 & 29.1412 & 28.9238 & 28.9389 & 29.3947 \\
\hline
DnCNN-S & 0.8580 & 0.8515 & 0.8670 & 0.8478 & 0.9032 & 0.8544 & 0.8425 & 0.8559 & 0.8514 & 0.7841 & 0.7949 & 0.8029 & 0.8428 \\
\hline
Proposed(Primary) & \bf 0.8662 & \bf 0.8589 & \bf 0.8729 & \bf 0.8604 & \bf 0.9116 & \bf 0.8584 & \bf 0.8459 & \bf 0.8669 & \bf 0.8795 & \bf 0.7978 & \bf 0.8022 & \bf 0.8189 & \bf 0.8533 \\
\hline
\end{tabular}
}
\end{center}
\caption{Performance comparison in terms of SSIM for ``Set12'' dataset in the Gaussian denoising task. The primary architecture was used.}
\label{Set12_SSIM_denoising}
\end{table*}
%\begin{table}[!hbt]

{\bf Dataset for NTIRE competition}: We used only 800 training dataset of DIV2K for each SISR problem. Instead of using cropped images, we cropped ($20\times 20$) patches randomly from a full image and performed data augmentation using image flipping, rotation, and downsampling with the corresponding scale factors of x2, x3, and x4 for each epoch. It helps to create more diverse patterns of images. For training dataset with bicubic x3 and x4 decimation, we used all bicubic x2,x3 and x4 images of DIV2K datasets together as a kind of data augmentation.

\begin{figure*}[!hbt]
\centerline{\includegraphics[width=0.88\linewidth]{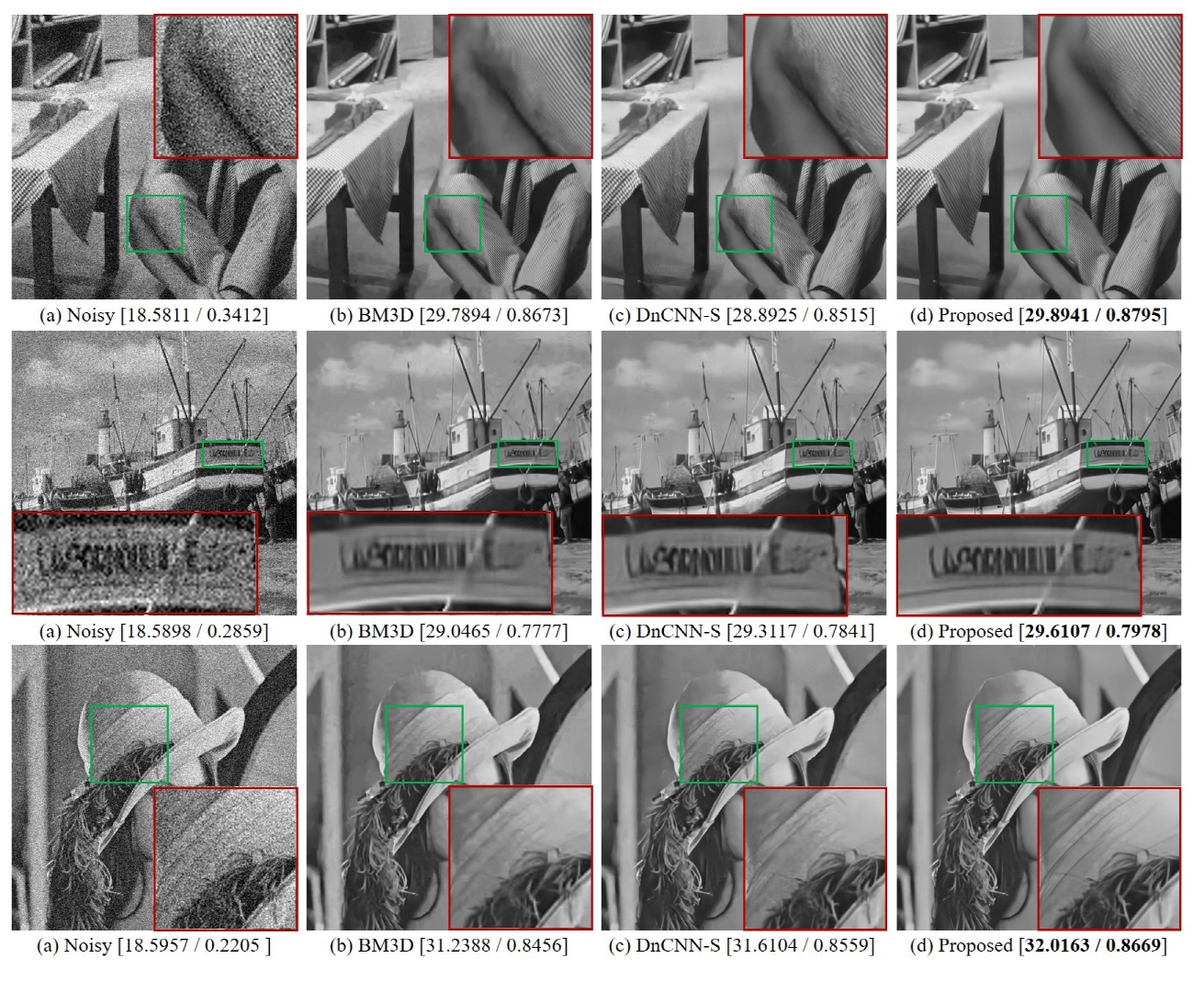}}
\caption{Denoising results of Barbara, Boats, and Lena images using various methods. [PSNR/SSIM] values are displayed.}
\label{fig:proposed_denoised_compareIMG}
\end{figure*}

\subsection{Network training}
{\bf Denoising network}:
The network parameters were initialized using the Xavier method \cite{he2015delving}. We used the regression loss across four wavelet subbands under $l_2$ penalty and the proposed network was trained by using the stochastic gradient descent (SGD). The regularization parameter ($\lambda$) was $0.0001$ and the momentum was 0.9. The learning rate was set from $10^{-1}$ to $10^{-4}$ which was reduced in log scale at each epoch. The mini-batch size for batch normalization was 32 where the images were selected randomly at every epoch \cite{ioffe2015batch}.
To use a high learning rate and guarantee a stable learning, we employed the gradient clipping technique \cite{kim2015accurate} so that the maximum and minimum values of the update parameter are bounded by the predefined range. These parameter settings were equally applied to all experiments of the image denoising. We used $40\times 40$ patch size and the network was trained using 53 epochs.
%{On the other hand, in super-resolution training, $20\times 20$ patch size and 30 epochs were used. We trained a single model network for multiple scale factors of two, three, and four.

The network was implemented using MatConvNet toolbox (beta.20) \cite{vedaldi2015matconvnet} in MATLAB 2015a environment (MathWorks, Natick). We used a GTX 1080 graphic processor and i7-4770 CPU (3.40GHz). The Gaussian denoising network took about two days for training.

{\bf Training for NTIRE competition}: 
We used $20\times 20$ patch size and trained the network for 150 epochs.
%Training method is similar to SISR of primary architecture except hyper parameter and method of cropping patch images. 
We used 64 mini-batch size and learning rate of $(0.1,0.00001)$ in log scale for 150 epochs with 0.05 gradient clipping factor. For each epoch, to train more various patterns, we used sub-epoch system that repeats forward and back propagation 512 times by randomly cropping patches from a full size image. For the bicubic x3 and x4 cases, we trained the network with the bicubic x2, x3, and x4 datasets together to increase the performance of x3 and x4 cases \cite{kim2015accurate}. Other hyper parameters are remained same with the denoising network. Using GTX 1080ti, the training of networks for the bicubic x2 and bicubic x3, x4 and unknown x2, x3, x4 datasets took almost six days, 21 days and seven days, respectively.

%\begin{figure*}[!hbt]
% \centerline{\includegraphics[width=0.76\linewidth]{fig/denoising_result_v4.png}} %0.77
% \caption{Denoising results of cropped 'Boats' and 'Lena' images using various methods for $\sigma=30$. [PSNR/SSIM] values are displayed. The proposed network is primary architecture.} %'Barbara'
% \label{fig:proposed_denoised_compareIMG}
%\end{figure*}
%

\section{Results}

\subsection{Persistent homology results}

To show the correlation between the network performance and manifold simplification, we compared the topology of the input and the label manifolds in both image and wavelet domains. 
The results in the supplementary material clearly showed that  feature space mappings  provide simpler data manifolds.
Specifically, the proposed denoising and super-resolution algorithms can be benefited from   simpler \emph{ input} manifold  from a feature space mapping using wavelet transform 
as well as additional simpler {\em label} manifold from residual learning.

\subsection{Experimental Results}

{\bf Denoising:}
For the quantitative comparison of the denoising performance, we used the objective measures such as the peak signal to noise ratio (PSNR) and the structural similarity index measure (SSIM) \cite{wang2004image}.
Table~\ref{Set12_PSNR_denoising},~\ref{Set12_SSIM_denoising} show that the proposed network outperforms the state-of-the-art denoising methods in terms of PSNR and SSIM for all Set12 images. Especially, in the patterned images such as Barbara and House, we attained better performance than using BM3D in terms of PSNR (0.1dB and 0.8dB, respectively).
Fig.~\ref{fig:proposed_denoised_compareIMG} shows the denoising examples in various images. The %House
proposed methods showed the best visual quality especially in the edge regions.
Moreover, as shown in Table~\ref{BSD68_denoising}, the proposed method showed superior results to the state-of-the-art approaches in the experiments with BSD68 dataset
which contains diverse patterned images.
For 512x512 image size, the proposed network took only 0.157 seconds even with the current MATLAB implementation. This is comparable or even better than the existing approaches.

\begin{figure}[!hbt]
\centerline{\includegraphics[width=0.6\linewidth]{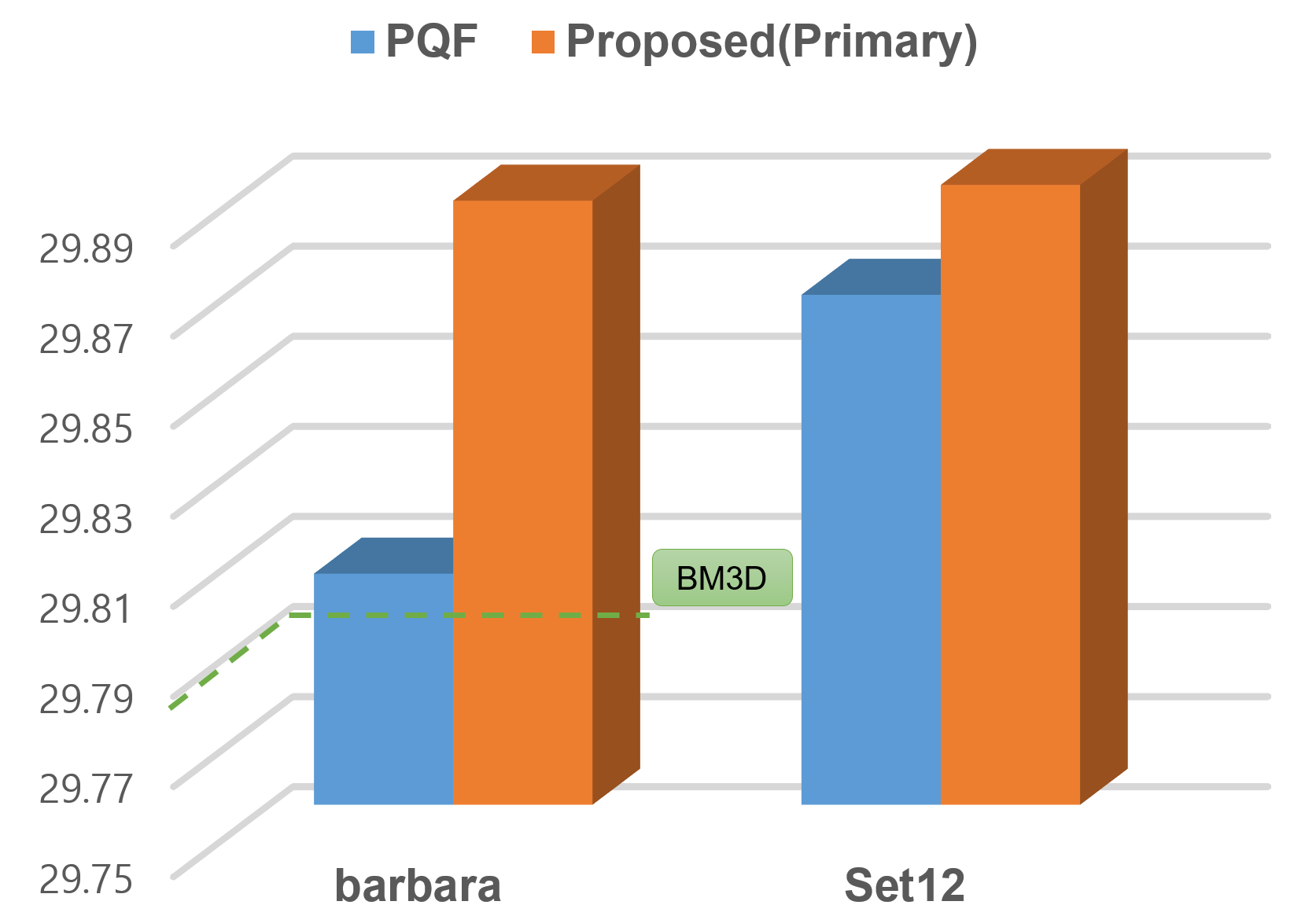}}
\caption{Importance of the wavelet transform for manifold simplification. Here, the Gaussian denoising algorithms with $\sigma$ = 30. Barbara and Set12 images were used. In the case of BM3D, Barbara image was used for comparison.}
\label{fig:comparison_of_waveletdecomposition}
\end{figure}

To further demonstrate the importance of the wavelet decomposition, additional comparative studies with the baseline network were performed. %of Fig.~\ref{fig:PQF_network}
Here, an input image is decomposed to four channels using so-called {\em polyphase quadrature filter (PQF)} bank \cite{vetterli1995wavelets}.
Specifically, the PQF just splits an input image into four equidistant sub-bands with distinct horizontal and vertical offset without wavelet filtering. Therefore, it is equivalent to the sub-pixel shuffling scheme in \cite{shi2016real} except that 
input image is first interpolated.
Accordingly, the networks using PQF have the exactly same architecture except the input and label images so that we can investigate the effect of the wavelet transform.
Fig.~\ref{fig:comparison_of_waveletdecomposition} clearly shows that the wavelet transform can improve the performance compared to the baseline network.

\begin{table}[t]
\begin{center}
%\resizebox{\columnwidth}{!}{%
\scalebox{0.73}{
\begin{tabular}{|c|c|c|c|c|} 
\hline
%\multirow{4}{1}{XXX}
Noise ($\sigma$) & BM3D & DnCNN-S & Proposed \\ %[0.5ex] 
\hline\hline
15 & 31.0761/0.8722 & 31.7202/0.8901 & \bf 31.8607/0.8941 \\ %[0.5ex] 
\hline 
30 & 27.7492/0.7735 & 28.3324/0.8003 & \bf 28.5599/0.8092\\ %[0.5ex]
\hline 
50 & 25.6103/0.6868 & 26.2275/0.7163 & \bf 26.3577/0.7270 \\ %[0.5ex] 
\hline 

\end{tabular}
}
\end{center}
\caption{Performance comparison in terms of PSNR/SSIM for ``BSD68'' dataset in the Gaussian denoising task.}
\label{BSD68_denoising}
\end{table}

\begin{table}[!hbt]
\begin{center}
%\resizebox{\columnwidth}{!}{% 0.73
\scalebox{0.72}{
\begin{tabular}{|c|c|c|c|c|c|} 
\hline
Dataset (scale) & VDSR & DnCNN-3 & Proposed-P & Proposed \\ %[0.5ex] 
\hline\hline
Set5 (2)& 37.53/0.9586 & 37.53/0.9582 	& 37.57/0.9586				& {\bf 38.06/0.9602} \\ %[0.5ex] 
\hline 
Set14 (2)& 33.03/0.9124 & 33.08/0.9126 & 33.09/0.9129					& {\bf 34.04/0.9205} \\ %[0.5ex] 
\hline 
BSD100 (2)& 31.90/0.8960 & 31.90/0.8956 & 	31.92/0.8965				& {\bf 32.26/0.9006} \\ %[0.5ex] 
\hline 
Urban100 (2)& 30.76/0.9140 & 30.75/0.9134 & 30.96/0.9169					& {\bf 32.63/0.9330} \\ %[0.5ex] 
\hline\hline 
Average (2)& 33.30/0.9202 & 33.31/0.9199 & 33.39/0.9212 					& {\bf 34.25/0.9286} \\ %[0.5ex] 
\hline\hline 
Set5 (3)& 33.66/0.9213 & 33.73/0.9212 & 33.86/0.9228				& {\bf 34.45/0.9272} \\ %[0.5ex] 
\hline 
Set14 (3)& 29.77/0.8314 & 29.83/0.8321 & 29.88/0.8331					& {\bf 30.56/0.8450} \\ %[0.5ex] 
\hline 
BSD100 (3)& 28.82/0.7976 & 28.84/0.7976 & 28.86/0.7987					& {\bf 29.18/0.8071} \\ %[0.5ex] 
\hline 
Urban100 (3)& 27.14/0.8279 & 27.15/0.8272 & 27.28/0.8334					& {\bf 28.50/0.8587} \\ %[0.5ex] 
\hline\hline 
Average (3)& 29.85/0.8445 & 29.89/0.8445	& 29.97/0.8470				 & {\bf 30.67/0.8595} \\ %[0.5ex] 
\hline\hline 
Set5 (4)& 31.35/0.8838 & 31.40/0.8837 	& 31.52/0.8864				& {\bf 32.23/0.8952} \\ %[0.5ex] 
\hline 
Set14 (4)& 28.01/0.7674 & 28.07/0.7681		& 28.11/0.7699			 & {\bf 28.80/0.7856} \\ %[0.5ex] 
\hline 
BSD100 (4)& 27.29/0.7251 & 27.29/0.7247 	& 27.32/0.7266				& {\bf 27.66/0.7380 }\\ %[0.5ex] 
\hline 
Urban100 (4)& 25.18/0.7524 & 25.21/0.7518 & 25.36/0.7614					& {\bf 26.42/0.7940 }\\ %[0.5ex] 
\hline\hline 
Average (4)& 27.95/0.7821 & 27.99/0.7820 	& 28.08/0.7861				& {\bf 28.78/0.8032 }\\ %[0.5ex] 
\hline

\end{tabular}
}
\end{center}
\caption{Performance comparison in terms of luminance PSNR/SSIM for various datasets in SISR tasks. The VDSR, DnCNN-3, Proposed-P(primary) networks are 291 dataset\cite{kim2015accurate} luminance-trained network whereas the proposed network is trained using RGB of DIV2K dataset.
For fair comparison, restored RGB was used to calculate the luminance PSNR/SSIM values. }
\label{Result_of_SuperResolution}
\end{table}

\begin{figure*}[t]
\centerline{\includegraphics[width=0.98\linewidth]{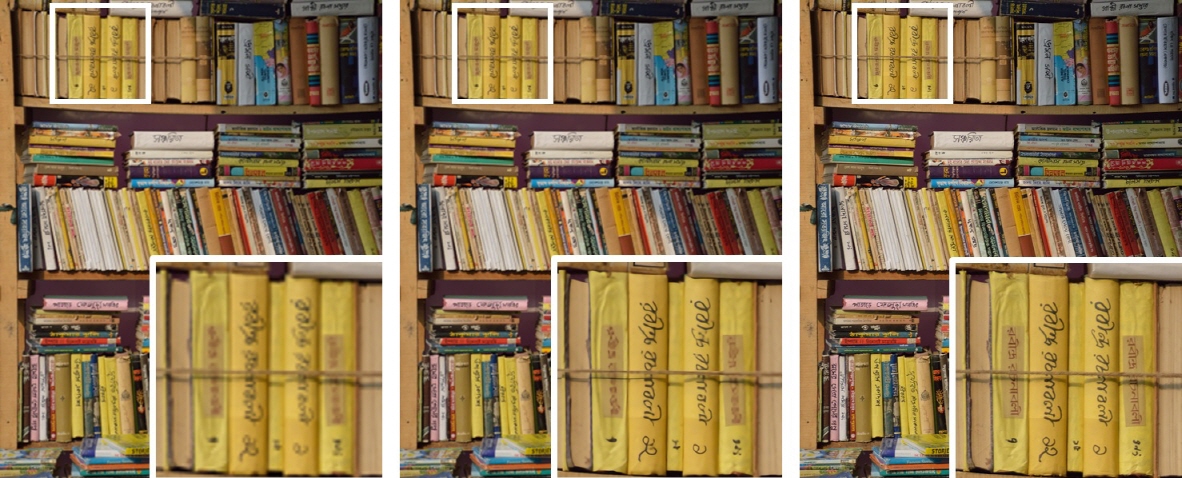}}
\caption{Performance comparison of SISR at scale factor of 4 of bicubic downsampling. The proposed network is RGB based competition network. Left : input, Center : restoration result, Right : label. }
\label{fig:SISR_compare_NTIRE_Bicubic_v1}
\end{figure*}

\begin{figure*}[t]
\centerline{\includegraphics[width=0.98\linewidth]{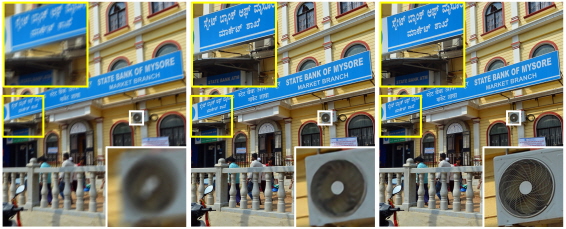}}
\caption{Performance comparison of SISR at scale factor of 4 of {\em unknown} downsampling. The proposed network is RGB based competition network. Left : input, Center : restoration result, Right : label. }
\label{fig:SISR_compare_NTIRE_unknown_v5}
\end{figure*}

{\bf NTIRE SISR competition:} 
Our proposed networks were ranked third in NTIRE SISR competition.
In particular, our networks were competitive in both performance and speed. More specifically, compared to the 14-67 seconds computational time by the top ranked groups' results,
our computational time was only 4-5 seconds for each frame.
%Moreover, compared to the unknown% decimation dataset, 
Since the wavelet transform is effective in reducing the manifold complexity compared to the sub-pixel shuffling scheme, in terms of dataset specific
ranking, our ranking  with the bicubic SISR dataset was  better than with the unknown decimation dataset where we just exploited the
manifold simplification from the residual learning.

%In Fig.~\ref{fig:NTIRE_compare1}, 
Figs.~\ref{fig:SISR_compare_NTIRE_Bicubic_v1}, \ref{fig:SISR_compare_NTIRE_unknown_v5}
clearly show the performance of our network in various SISR problems. We also confirmed that our network outperforms the existing state-of-the-art CNN approaches for various dataset in Table~\ref{Result_of_SuperResolution}. In particular, the proposed methods exhibited outstanding performance in the edge areas. We provide more comparative examples in the supplementary material.

\section{Conclusion}
In this paper, we proposed a feature space deep residual learning algorithm that outperforms the existing residual learning approaches. %as a novel deep network design criterion. % using manifold simplification, inspired by the 
In particular, using persistent homology analysis,
we showed that
% Moreover, based on the observation that 
the wavelet transform and/or residual learning results in simpler data manifold. This finding as well as the experimental results of Gaussian denoising and NTIRE SISR competition results
confirmed that the proposed approach is quite competitive in terms of performance and speed.
%outperforms the existing state-of-the art approaches.
Moreover, we believe that the persistent homology-guided manifold simplification 
provides a novel design tool for general deep learning networks.
% in both denoising and image super-resolution problems.
% Moreover, the performance of the propring approaches across all images with many patterns.

\section{Acknowledgement}

This work is supported by Korea Science and Engineering Foundation, Grant number
NRF-2013M3A9B2076548.

%\clearpage
%
%{\small
%\bibliographystyle{ieee}
%\bibliography{egbib_woong}
%}

%----------------------------------------------------------------------------------------------------------------------------------------
%\begin{document}

%%%%%%%%% TITLE
%\title{(Supplementary Material) Beyond Deep Residual Learning for Image Restoration: \\ Persistent Homology-Guided Manifold Simplification }

%\author{Woong Bae, Jaejun Yoo, and Jong Chul Ye\\
%Bio Imaging Signal Processing Lab\\
%Korea Ad. Inst. of Science \& Technology (KAIST) \\
%291 Daehak-ro, Yuseong-gu, Daejeon 34141, Korea
%\\
%{\tt\small \{iorism,jaejun2004,jong.ye\}@kaist.ac.kr}\\
%{\small * denotes co-first authors}% For a paper whose authors are all at the same institution,
%% omit the following lines up until the closing ``}''.
%% Additional authors and addresses can be added with ``\and'',
%% just like the second author.
%% To save space, use either the email address or home page, not both
%}

\maketitle
%\thispagestyle{empty}

%%%%%%%%% BODY TEXT

%-Appendix ---------------------------------------------------------------------------------------------------------------------------------------
\clearpage
\section{Appendix}

\subsection{Persistent homology results}
\label{sec:ph_result}

For the denoising task, total 4500 numbers of $40\times 40$ patches with four components of the wavelet transformed data and $80\times 80$ patches of the image domain data were set to a set of point clouds in $\Rd^{6400}$ vector space of both image and wavelet domain manifolds. Note that the patch size of the image domain is doubled to match the size of the receptive field.
%These point cloud were generated to investigate the input and label manifolds. 
%The residuals from both original and wavelet domain images were investigated as the label manifolds.
For the super-resolution experiments, a set of $20\times 20$ patches was cropped to generate a point cloud in both bicubic and unknown datasets. Similarly, $40\times 40$ image patch was used for image domain approaches.

To investigate the topology of the dataset, a metric should be defined.
Since we used the $l_2$ regression error as the loss, the natural choice should be $l_2$ distance, i.e. $d_2(X_i, X_j) = \|X_i-X_j\|_2$.
However, care need to be taken for the input dataset, because the data should pass through the batch-normalization
which changes their mean and variance. 
Note the distance metric that is invariant under batch normalization is the correlation-based metric:
$d_{corr}(X_i, X_j) = \sqrt{1-corr(X_i, X_j)},$
where $corr(X_i, X_j)$ denotes the normalized Pearson's correlation between $X_i$ and $X_j$.
Therefore, we used the $d_{corr}$ as the metric for the input space, whereas $d_2$ was used as a metric for the label space.
%Note that this metric is invariant under batch normalization.

\begin{figure}[!hbt]
\centerline{\includegraphics[width=0.99\linewidth]{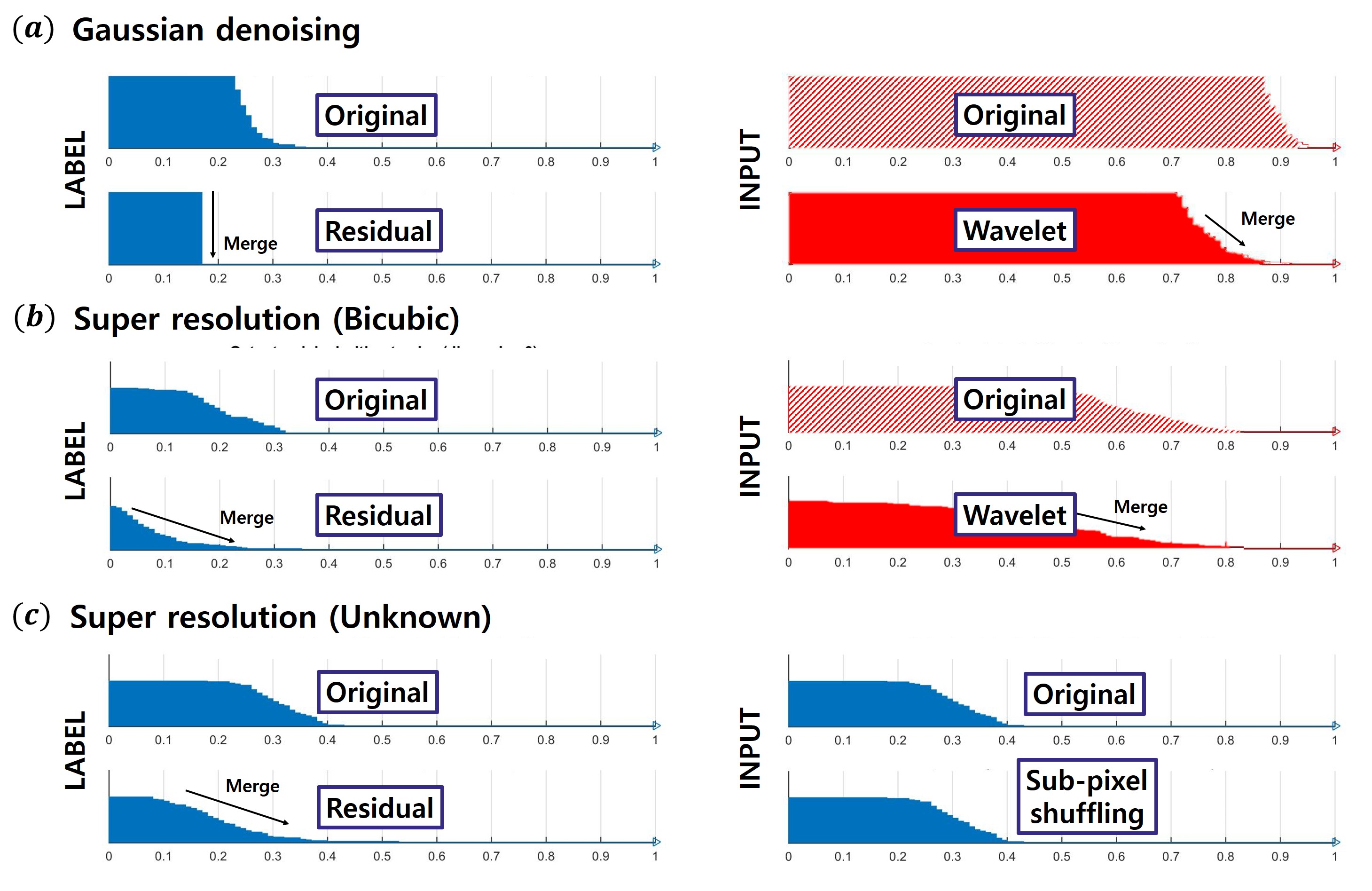}}
\caption{ (Left) Persistent homology analysis of {\em label} manifold. Zero dimensional barcodes of the label manifolds between the original and residual images are compared for (a) Gaussian denoising , (b) super-resolution (bicubic decimation), and (c) super-resolution (unknown decimation) tasks. 
(Right)  Persistent homology analysis of {\em input} manifold. Zero dimensional barcodes of the input manifolds between the original and feature space mappings are compared for (a) Gaussian denoising, and (b) super-resolution (bicubic decimation),  and (c) super-resolution (unknown decimation) tasks. }
\label{fig:ori_resid}
\end{figure}

To show that the advantages of residual learning comes from the simpler topology of residual labels,
we first calculated the barcodes of the original images and residuals. 
We calculated Betti numbers and barcodes using a toolbox called JAVAPLEX (http://appliedtopology.github.io/ javaplex/). 
The barcodes in the left column of the Fig.~\ref{fig:ori_resid} clearly show that
the topology of the label manifold composed of residual image patches is much simpler. The point clouds of the residual image patches merged earlier than that of the original ones.
Note that the residual manifolds of both image and wavelet domains have an identical topological complexity in the image domain residual due to the orthogonal Haar wavelet transform.
 The barcodes of residual manifolds in Gaussian denoising case showed rapid drops at $\epsilon=0.17 $, which is related to the noise standard deviation.
%To perform quantitative evaluation for these findings, we 
%calculated persistence landscape. Total 500 patches were randomly drawn from each dataset. Then, the topological difference was tested by using two-sample $t$-test with 100 trials of landscapes per each group. 
%%The distributions of persistence landscapes are shown in Fig.\ref{fig:ori_resid}. Again, the residual manifolds showed significantly lower persistence landscape values ($p\ll 0.001$). %
These results imply that the label manifold for residual  has much simpler topology than that of original one.
% as it did in the spherical example in Fig.\ref{fig:topo_exp}(a). 
We believe that this is the main 
reason for the success of deep residual learning.

%\begin{figure}[!hbt]
%\centerline{\includegraphics[width=1\linewidth]{fig/persistent_homology_results_v2.jpg}}
%\caption{ Simpler {\em transformed input} manifold. Zero dimensional barcodes of the input (red) versus their residual labels (blue) are compared between the original (purple) and wavelet (orange) domains for both (a) Gaussian denoising, and (b) super-resolution(Bicubic) tasks. The distributions of persistence landscape for both original and residual images are compared.}
%\label{fig:topo}
%\end{figure}

Next, we investigated the advantages of wavelet transform which further reduces the topological complexity of the input spaces in both Gaussian denoising and super-resolution datasets (the right column of Fig.~\ref{fig:ori_resid}). While the residual manifolds of both image and wavelet domains had a very similar topological complexity represented by barcodes, the input manifold of the wavelet transformed images had simpler topology than the original images. 
%These trends were demonstrated by the landscape analysis results whose distribution of input space showed a significant difference while the differences between two residual manifolds were not statistically significant. 
However, the sub-pixel shuffling in the unknown decimation case does not change the 
manifold structure because it  merely shuffles the order of the pixels. Accordingly,  the topology of the input manifold does not change, but  the residual learning in sub-pixel shuffling domain reduces the complexity
of the label manifold.

%We believe that the first 
Accordingly, we expect that the proposed denoising and super-resolution algorithms are benefited from  the simpler \emph{ input} manifolds from feature space mapping
as well as additional simpler {\em label} manifold from residual learning.
%whereas the proposed super-resolution algorithm is benefited from simplified topology in both input and label manifolds.

%
%\begin{figure}[!t]
% \centerline{\includegraphics[width=0.9\linewidth]{fig/Original_vs_Residual_v2.png}}
% \caption{Topological characteristic of original and residual manifolds in (a) Gaussian denoising and (b) super-resolution. 
% The $\beta_0$ of original and residual images and the distributions of \eqref{eq:lambda} were calculated.}
% \label{fig:topo}
%\end{figure}

% \vspace*{-1cm}

%\begin{figure*}[!hbt]
%\centerline{\includegraphics[width=0.98\linewidth]{fig/NTIRE_unknown_x4_v2.png}}
%\caption{Performance comparison of SISR at scale factor of 4 of unknown downsampling. The proposed network is RGB based competition network. Left : input, Center : restoration result, Right : Label. }
%\label{fig:SISR_compare_NTIRE_unknown_v2}
%\end{figure*}

\begin{figure*}[!hbt]
\centerline{\includegraphics[width=0.99\linewidth]{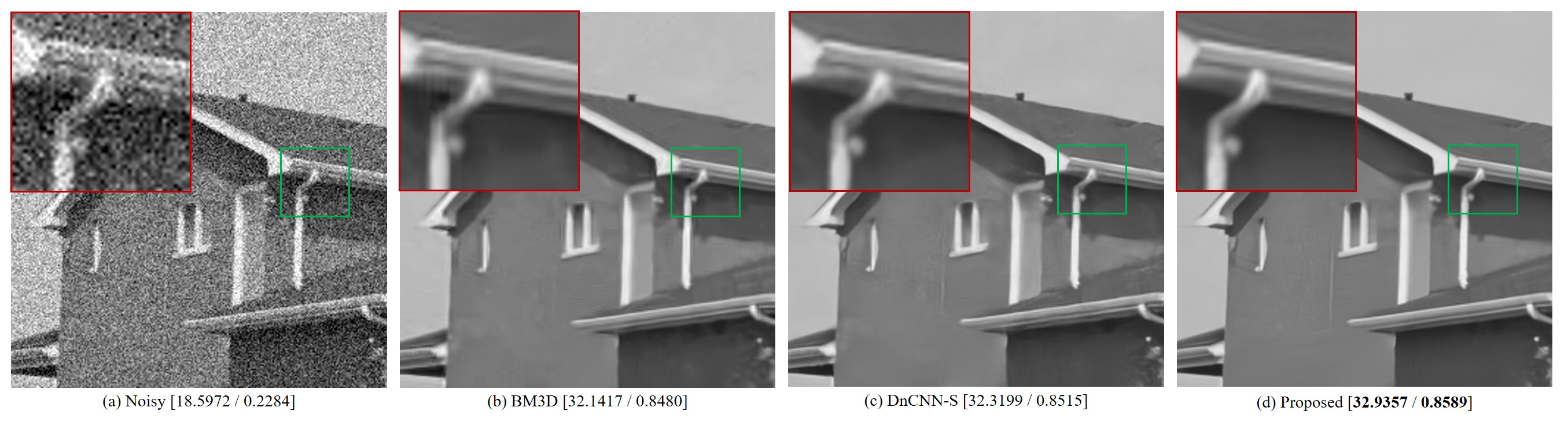}} 
\caption{Denoising performance comparison. PSNR/SSIM values are displayed. Gaussian noise with $\sigma$=30 were added. .}
\label{fig:camera}
\end{figure*}
\begin{figure*}[!hbt]

\centerline{\includegraphics[width=0.99\linewidth]{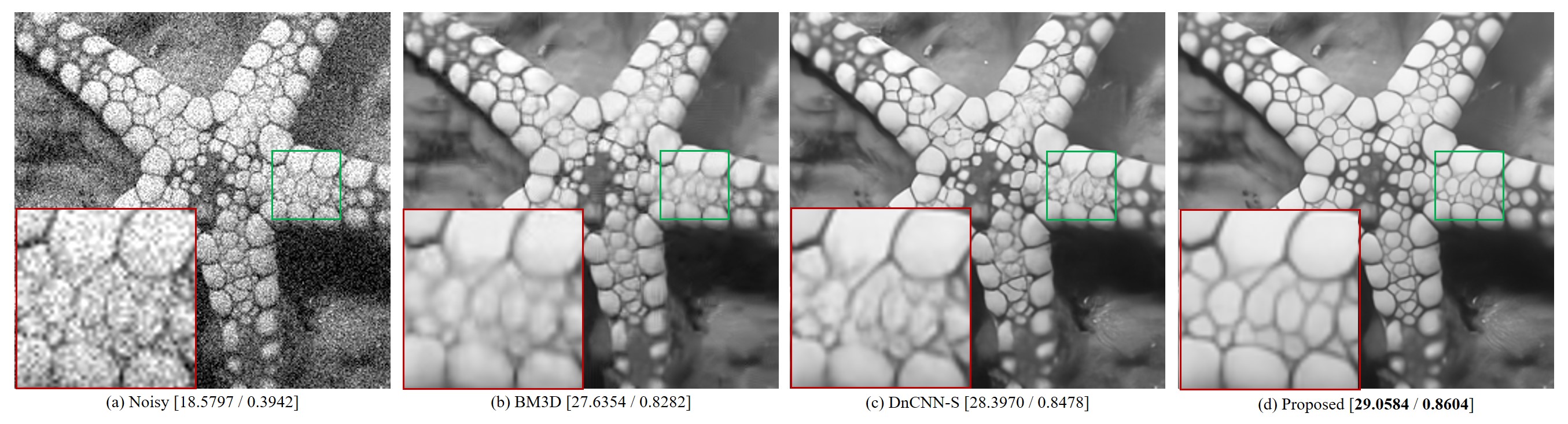}} 
\caption{Denoising performance comparison. PSNR/SSIM values are displayed. Gaussian noise with $\sigma$=30 were added. .}
\label{fig:camera2}
\end{figure*}

\begin{figure*}[!hbt]
\centerline{\includegraphics[width=0.99\linewidth]{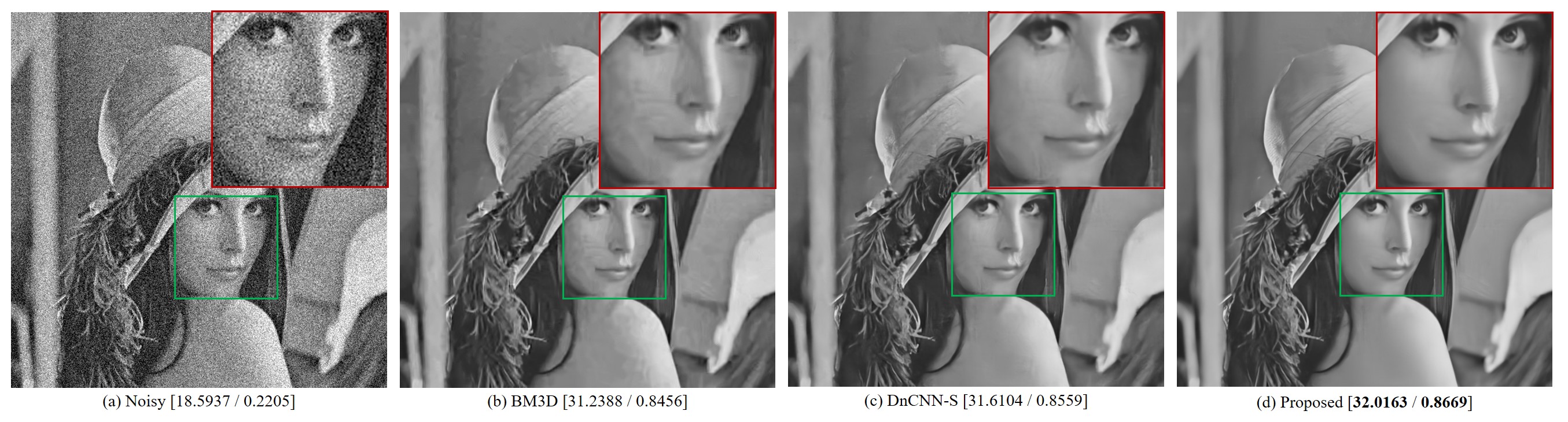}} 
\caption{Denoising performance comparison. PSNR/SSIM values are displayed. Gaussian noise with $\sigma$=30 were added. .}
\label{fig:lena}
\end{figure*}
\vspace*{1cm}
\subsection{Additional Results from Gaussian denoising and NTIRE SISR competition}
\label{sec:ph_results_supplementary_GD_SISR_NTIRE}
In this section, we provide more results for Gaussian denoising and SISR reconstruction from NTIRE competition dataset.
%\clearpage
%
%\begin{figure}[!hbt]
% \centerline{\includegraphics[width=0.90\linewidth]{fig/barbara_v2.png}} 
% \caption{Denoising performance comparison. PSNR/SSIM values are displayed. Gaussian noise with $\sigma$=30 were added. .}
% \label{fig:camera}
%\end{figure}

%
%\subsection{SISR based on RGB for NTIRE competition networks}
%\label{sec:ph_result_SISR_NTIRE}
%
%In this section, we provide more SISR results from NTIRE competition.

\begin{figure*}[!hbt]
\centerline{\includegraphics[width=0.98\linewidth]{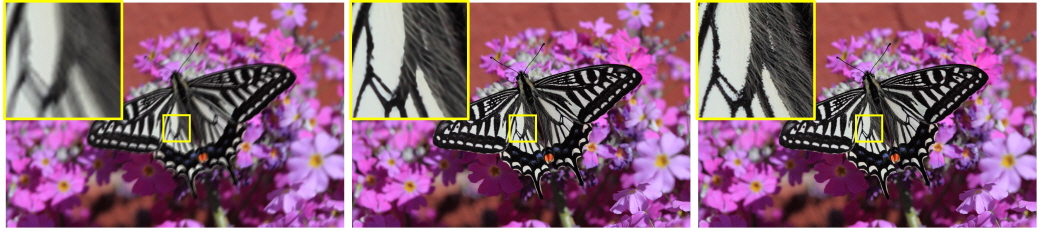}}
\caption{Performance comparison of SISR at scale factor of 4 of unknown downsampling. The proposed network is RGB based network. Left : input, Center : restoration result, Right : label. }
\label{fig:SISR_compare_NTIRE_unknown_v1}
\end{figure*}

\begin{figure*}[!hbt]
\centerline{\includegraphics[width=0.98\linewidth]{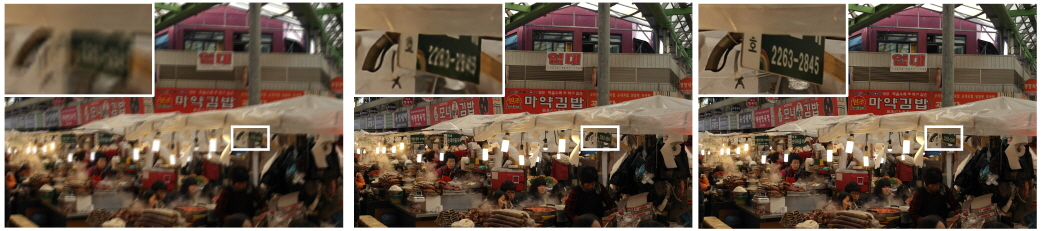}}
\caption{Performance comparison of SISR at scale factor of 4 of unknown downsampling. The proposed network is RGB based network. Left : input, Center : restoration result, Right : label. }
\label{fig:SISR_compare_NTIRE_unknown_v4}
\end{figure*}

\begin{figure*}[!hbt]
\centerline{\includegraphics[width=0.98\linewidth]{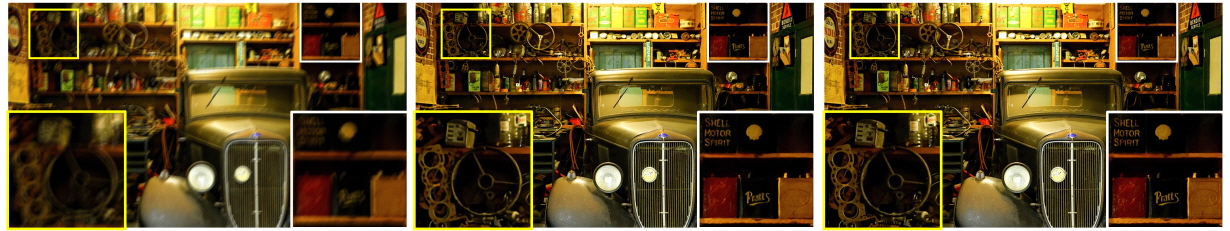}}
\caption{Performance comparison of SISR at scale factor of 4 of unknown downsampling. The proposed network is RGB based network. Left : input, Center : restoration result, Right : label. }
\label{fig:SISR_compare_NTIRE_unknown_v6}
\end{figure*}

\begin{figure*}[!hbt]
\centerline{\includegraphics[width=0.98\linewidth]{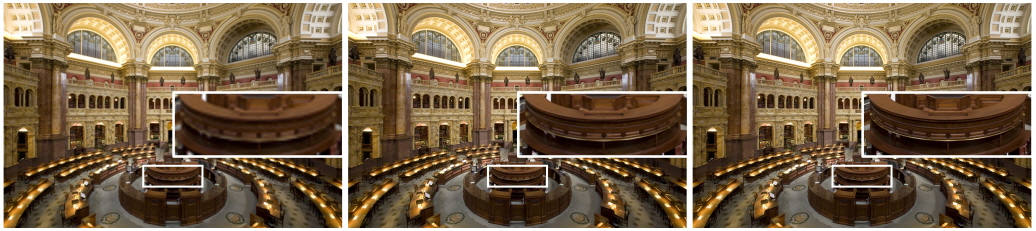}}
\caption{Performance comparison of SISR at scale factor of 4 of bicubic downsampling. The proposed network is RGB based network. Left : input, Center : restoration result, Right : label. }
\label{fig:SISR_compare_NTIRE_bicubic_v2}
\end{figure*}

\begin{figure*}[!hbt]
\centerline{\includegraphics[width=0.98\linewidth]{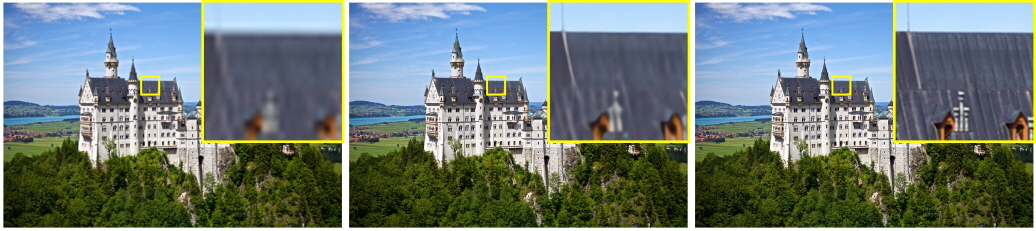}}
\caption{Performance comparison of SISR at scale factor of 4 of bicubic downsampling. The proposed network is RGB based network. Left : input, Center : restoration result, Right : label. }
\label{fig:SISR_compare_NTIRE_bicubic_v3}
\end{figure*}

\begin{figure*}[!hbt]
\centerline{\includegraphics[width=0.98\linewidth]{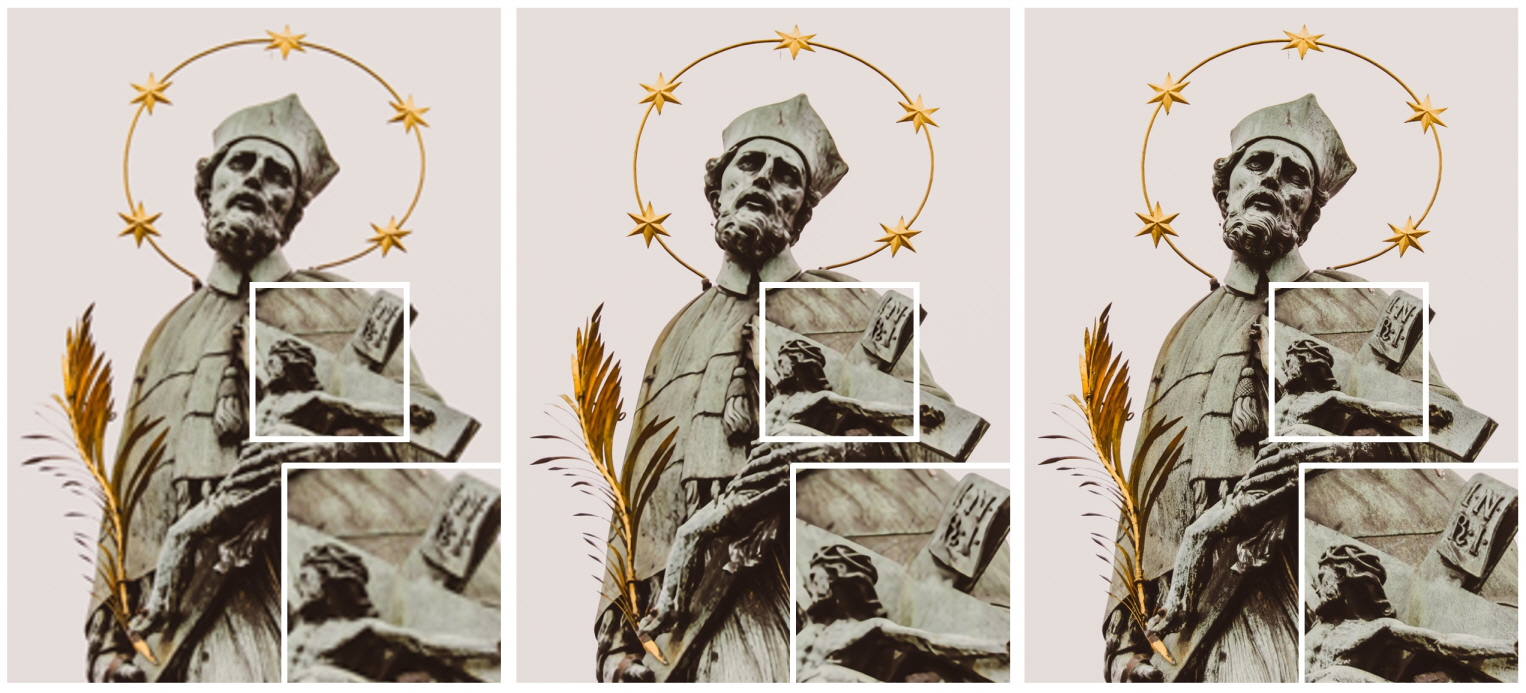}}
\caption{Performance comparison of SISR at scale factor of 4 of bicubic downsampling. The proposed network is RGB based network. Left : input, Center : restoration result, Right : label. }
\label{fig:SISR_compare_NTIRE_bicubic_v4}
\end{figure*}

\begin{figure*}[!hbt]
\centerline{\includegraphics[width=0.98\linewidth]{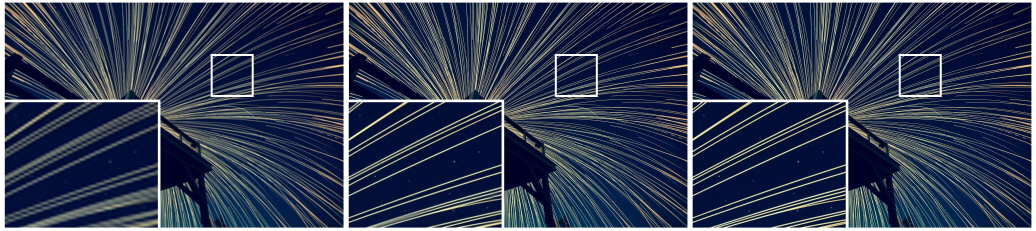}}
\caption{Performance comparison of SISR at scale factor of 4 of bicubic downsampling. The proposed network is RGB based network. Left : input, Center : restoration result, Right : label. }
\label{fig:SISR_compare_NTIRE_bicubic_v5}
\end{figure*}

%%-------------------------------------------------------------------------

%\end{document}

\end{document}